# Procedural Urban Environments for FPS Games

Jan Kruse
Auckland University of Technology
Private Bag 92006, Wellesley Street
Auckland 1142, New Zealand
+64 9 921 9999
jan.kruse@aut.ac.nz

Ricardo Sosa
Auckland University of Technology
Private Bag 92006, Wellesley Street
Auckland 1142, New Zealand
+64 9 921 9999
ricardo.sosa@aut.ac.nz

Andy M. Connor
Auckland University of Technology
Private Bag 92006, Wellesley Street
Auckland 1142, New Zealand
+64 9 921 9999
andrew.connor@aut.ac.nz

## ABSTRACT
This paper presents a novel approach to procedural generation of urban maps for First Person Shooter (FPS) games. A multi-agent evolutionary system is employed to place streets, buildings and other items inside the Unity3D game engine, resulting in playable video game levels. A computational agent is trained using machine learning techniques to capture the intent of the game designer as part of the multi-agent system, and to enable a semi-automated aesthetic selection for the underlying genetic algorithm.

## CCS Concepts
• Software and its engineering➝Software organization and properties➝ Contextual software domains➝Virtual worlds software➝Interactive games

• Applied computing➝Computers in other domains➝ Personal computers and PC applications➝ Computer games

## 1. INTRODUCTION
First Person Shooter (FPS) games have been one of the most successful and fastest growing genres in the computer games industry over the past decade [1]. Titles such as *Battlefield, Insurgency, Halo, Call of Duty* and many others sell millions of copies every year. Creating engaging playable maps for FPS games is a challenging and laborious task [2] which is usually achieved manually which ensures that the level is both playable and matches the intent of the designer. Urban environments are particularly challenging as they are characterized by high complexity. The large number of individual items such as houses, streets, cars or similar obstacles and smaller props pose huge challenges for game level designers. This paper outlines a novel semi-automated approach to urban map design, which draws on Interactive Evolutionary Computation and Autonomous Agents to produce fully playable FPS game levels. The approach has the potential to significantly reduce the degree of manual labour, making the design process more efficient.

## 2. BACKGROUND AND RELATED WORK
Procedural level creation is a popular application for creative computational systems [3] and has been used since the 1980s to produce generative content for video games [4]. It gained new momentum in recent years due to the availability of computational resources in form of faster CPUs and general-purpose computing on GPUs [5]. However, most approaches to procedural content generation are designed to replace the human game designer. Such approaches offer both benefits and disadvantages typical of applying Artificial Intelligence (AI) approaches to cognitive task. AI approaches can work much faster than a human level designer, and are in some cases able to explore the design space automatically to find levels with desirable qualities. But they aren't able to capture the human creativity that produces the most interesting level designs, and they are usually very specific to their particular domain [6].

### 2.1 Evolving FPS Maps
In this paper procedural or generative design denotes the automatic or semi-automatic generation of game levels, including the terrain, position of buildings, streets and other items. Instead of manually placing game assets into the map, significant parts of this task are automated. While there are some similarities to work conducted for instance by Cardamone et al. [7] in context of FPS content generation, and also by Cook and Colton [8] for general procedural 3D game design, this approach differs in that it heavily considers the (human) game designers intent and feeds this deliberate choice back into the automated selection process of the evolutionary algorithm. Our approach is not about getting computer to produce content or whole game procedurally, but more an attempt to support the game designer in an otherwise very costly and time consuming process. Currently, as a result of high design and development costs, FPS games offer only a few hours of single-player gameplay and oftentimes just a small number of multiplayer maps [7]. This paper offers a semi-automated approach that should allow level designers to produce a higher number of maps at significantly reduced costs which has the potential to boost the popularity of a new title as a means to commercial success.

Another important benefit of the proposed approach is the reduction of game testing costs by use of automated content evaluators in addition to pure content generators. The algorithm allows simple integration of additional agents, which provides flexible approaches to design and evaluation. Testing levels to fix issues around playability, technical flaws, enjoyableness and fair balance in case of multiplayer team-based gameplay is currently another significant issue for commercial game development. Some companies tend to release maps as open betas to get the gamer community to evaluate the maps and identify technical and ludic issues before the official release of the game or game extension. Other companies rely on "free to play" models to test the base content of the game before selling additional items commercially. Examples include *Battlefield Community Testing Environment* (free beta play), *Team Fortress 2* (free to play) and the recently announced *Rainbow Six Siege* ('closed' beta with access for anyone with an NVidia graphics card). Eventually, this work will also lead to reduced cost of map testing by use of additional evaluation agents, resulting in cost effective generation and evaluation of game content.

Evolving FPS game levels, also often called FPS maps, is not an entirely novel idea. Previous work includes Cardamone et al. [7] who also based the map generation on evolutionary algorithms, with the main difference that their work used a reasonably simple mathematical fitness function based on the average fighting time of the player plus the free space on the map. The fitness function is not only purely theory-driven rather than modelled on data-driven player capabilities and preferences, but it also reflects a purely player-oriented approach, which excludes the game designer's intent. Here the computational agent is the single designing entity. Our approach differs from this in that it considers the map designer and their preference. We use a design-driven instead of user-driven approach to map generation. Potential expansions of this idea is outlined below under Human-based Genetic Algorithms.

Another recent example for procedurally generated FPS maps is work conducted by Lanzi et al. [9] which in turn utilizes Search-based Procedural Content Generation [10]. Their work aims to evolve game levels that provide basic match balancing properties based on player skill and strategies. The content generator uses genetic algorithms and is not dissimilar to the level generation outlined above [7]. The novelty of our approach and main difference to their work lies in the addition of a multi-agent system and the ability to capture a human game designer's preferences.

## 2.2 Human-Based Genetic Algorithms

Human-based Genetic Algorithms (HBGA) are a variation of Interactive Genetic Algorithms [11]. The human user is replaced by a multi-agent system consisting of human and computational agents [12]. A generic example of a hybrid human and computational HBGA is shown in **Error! Reference source not found.**, where the human works in combination with an agent to perform selection from the population. An additional agent undertakes the recombination processes.

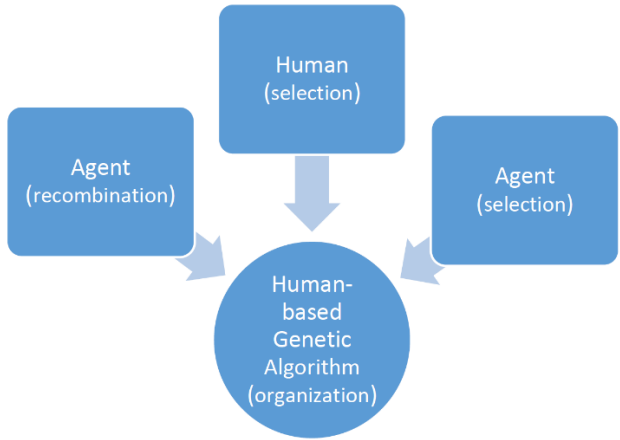

**Figure 1. Human-Based Genetic Algorithm.**

HBGA have been used to successfully counteract user fatigue, one of the implicit problems of Interactive Evolutionary Systems when applied to creative design tasks [13]. A machine learning classifier (J48) is trained using the designer's input over a number of GA generations to gauge the aesthetic preference and capture the design intent. An agent based on this classifier provides additional (computational) selections, which reduces the selection runs for the human in the HBGA. This approach has shown modest, but promising results in previous applications [13].

While it could be argued that calling an interactive genetic algorithm with additional agents a Human-based Genetic Algorithm is merely an academic exercise, the idea of a multi-agent system that feeds into a genetic algorithm goes beyond that. It poses a paradigm shift, where selection and recombination are separated and performed by different entities. This allows subsequently to add a variety of agents to the algorithm as we will show in the following sections of this paper. It is envisaged that such agents could focus on different aspects of the design, so potentially could include a diversity agent that drives the design towards novel layouts. This inclusion of multiple agents transforms canonical Interactive Evolutionary Computation as introduced by Takagi and Iba [14] from a purely generative algorithm into a flexible system for generation and evaluation. Therefore, using the concept of Human-based Genetic Algorithm proves to be a successful framework for procedural content for video games.

But the HBGA opens up additional possibilities, for instance running design intent and user preference concurrently as the driving forces behind map generation. The HBGA allows both design- and user-driven paradigms to be implemented by choosing selection agents accordingly. The multi-agent system could be made of both (human) designer agent for selection as well as (human) player agent for selection, both in additional to computational agents. Some games such as *Battlefield* or *Insurgency* allow users at the start of the round to select the map that is going to be played next. With HBGA map generation in mind, this pre-game 'vote' could be used to feed back into the HBGA to capture the player's intent and therefore increase the chance of more enjoyable gameplay for the target group.

## 3. URBAN FPS GENERATOR

We present an *Urban FPS Generator*, which is a multi-agent evolutionary system that integrates human and computational agents into the selection process of a genetic algorithm. It is capable of addressing different aspects of Computational Creativity, and in its current version targets placement of game assets to form playable FPS maps. Figure 2 shows a typical game level in Unity3D developed with the approach.

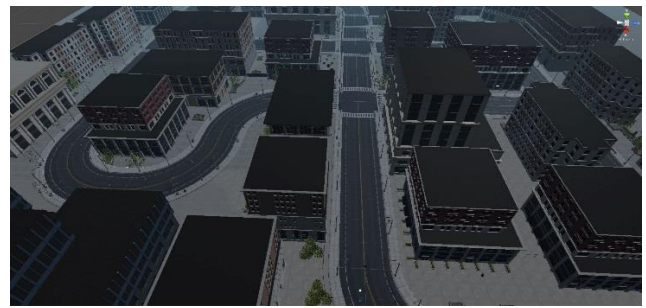

**Figure 2. Game Level in Unity3D.**

Urban FPS Generator extends the generic HBGA framework shown in Figure 1. The specific agents involved are shown in Figure 3.

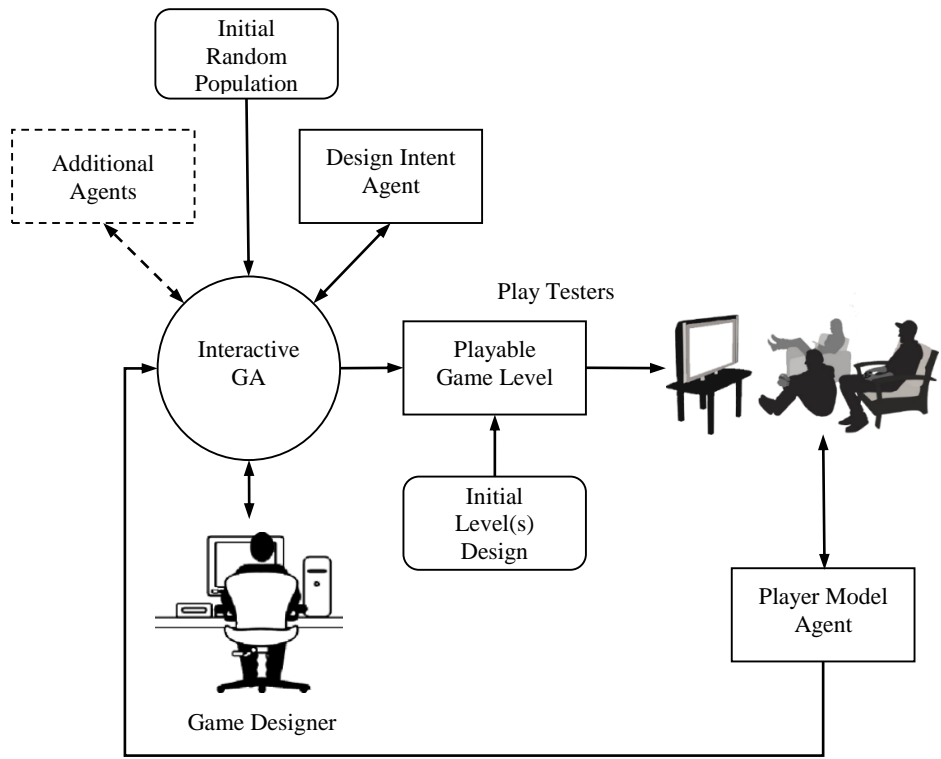

**Figure 3. Urban FPS Generator System Overview.**

The usage of the system would typically start with the manual design of a number of different game levels that are used in the development of the player model. These levels are played extensively by a group of suitable play testers and this gameplay is used to generate the player model. This development of the player model is performed "offline" from the game design process. When the game designer wishes to start the design of a new level, the interactive Genetic Algorithm generates a new initial random population. The creation of the next generation is enacted through a combination of the designer's choices and the input of the player model agent which evaluates the playability of each level. As the designer evolves the new level the design intent agent observes the designer's choices and learns about the goal of the designer, whether such a goal is even implicit. The design intent agent can accelerate the evolution of new levels by acting on the behalf of the designer and therefore reducing the fatigue associated with typical interactive evolutionary computation approaches. Whilst not yet developed, the agent architecture allows for additional agents to also influence the evolution of the new level. For example, a novelty agent may inject candidate solutions that are distinct from those generated by the designer and/or the design intent agent as a means to prevent premature convergence. The Human-Based Genetic Algorithm is therefore the combination of these agents with the interactive Genetic Algorithm and the designer.

Previous work on procedural generation of cityscapes [13] was implemented in Java, assisted by the core libraries of the *Processing* programming language as imported *jar* files to visualize the underlying map layouts. In contrast, *Urban FPS Generator* is written as an extension to Unity3D, a commercially available game development environment using C# (and optionally JavaScript). Unity3D offers cross-platform development for browser games, native PC platforms, game consoles and handheld devices in a highly versatile and extensible way. It is also a very efficient video game engine with support for the latest graphics options (DirectX 11 and OpenGL 4), which allows games to run at a playable framerate with minimal lag while the graphics are attractive and aesthetically pleasing. These attributes make it a good choice for FPS game research, and given that one of the aims of this project is a fully functional, distributable and simple to install application. This application is intended to serve as a platform for future research into player behaviour and agent systems, therefore Unity3D seems to be a better choice, superior to fully custom developments in Java.

The systems makes use of prefabs, which are prebuilt game assets inside the Unity3D project. Prefabs can be used as instances inside a game level (or scene). The current version of *Urban FPS Generator* uses 12 building prefabs, a range of street sections, a number of simple props such as chairs, tables, leaves, containers and plants, as well as 5 weapons from the UFPS package from the Unity3D asset store. This has allows the system to be suitably robust and have the ability to produce fully playable levels. One such level is shown in Figure 4.

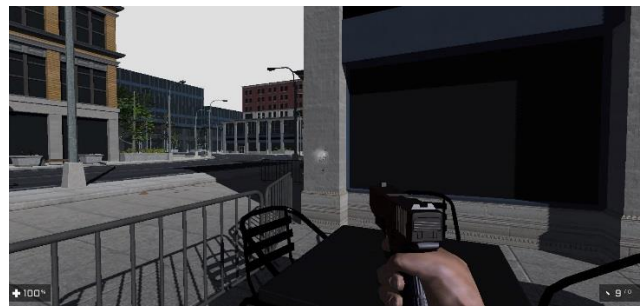

**Figure 4. Fully Playable FPS Map.**

We are currently working on extending this basic library of prefabs by adding additional items, but also even smaller building blocks. Our main intent is the generation of maps, but it is also possible to evolve the building blocks for these maps from even smaller units such as parts of buildings (windows, framing, doors, and wall pieces). Currently, *Urban FPS Generator* is not capable of creating interiors of multi-level buildings or complex maze-like structures such as interior office spaces or cluttered warehouses. Using partial prefabs to evolve buildings and other structures allows us to address these issues.

Further, we are building terrain support into the evolutionary system, which enables rugged environments such as mountain ranges and elevated city layouts to be part of the map generation. While this means an increase in parameters for the HBGA, the user is factually only interacting with the same interface and requires no additional skills or knowledge. At the same time, with an increase of the number of parameters, the need for additional iterations in the HBGA will rise and therefore compensation for user fatigue through learning computational agents is of even higher significance for the overall system performance.

### 3.1 User Interaction

The system presents a selection of 9 maps to the designer using the inbuilt UI system in Unity3D. The maps are rendered top-down views of three-dimensional game levels. The user selects the two most preferred maps, which are fed back into the evolutionary breeding pool. In parallel, the agent receives all 9 maps classified by the user selection as training data. The process is repeated for a maximum number of times or when terminated by the designer. The final map is used as the active game level for the FPS round.

### 3.2 Map Encoding

Unity3D supports 3D game levels of nearly any complexity, only limited by computer performance and creation time. We take a much simpler approach and limit the maps to currently only a single level. *Urban FPS Generator* is based on a two-dimensional matrix (512x512 units) which roughly follows Unity3D's terrain system with a few exceptions. We use zero elevation for the urban environments. Our street system also uses large sections of 25 units, which limits the number of streets and intersections in each dimension. This can easily be extended by increasing the base terrain size, but in order to keep computation times for the map generation to a minimum, we have opted for these rather small game levels. From experience with existing maps in AAA titles, this is not a disadvantage. *Battlefield* has an additional content package that explicitly focusses on close quarter, infantry only warfare. Given that *Urban FPS Generator* has no inbuilt functionality for vehicles, we believe this choice is sufficient and has no impact on playability or enjoyableness. This will be verified in future work.

The chromosomes of the maps are represented as a strongly typed list where all parameters such as building height, type of content (building, street, free space) and position of props (leaves, containers, obstacles, fences, plants) are held as numerical values. The system is able to produce environments where props are well placed in context, as shown in Figure 5.

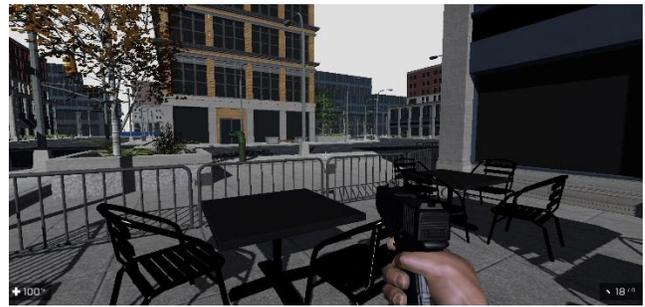

**Figure 5. Placement of Props.**

We use real coded values to counteract some of the known issues when dealing with variables in continuous domains [15]. Accordingly, crossover and mutation are performed on real values instead of binary strings.

### 3.3 Agents

The computational agents in our HBGA are currently based on J48 (open source C4.5 implementation) classifiers that feed into simple decision trees. Previous work [13] successfully utilised J48 as a classifier of designer intent, though we acknowledge that this might not be the ideal classifier for this type of machine learning application due to the inherent system noise based on the low number of training samples (user inputs) in relation to the number of attributes being classified. J48 has been acknowledged as problematic in this regard in other applications [16]. As further attributes are introduced to the classifier as *Urban FPS Generator* is extended is expected that J48 will reach a threshold level of performance. We are looking at alternatives in a different context and will replace this algorithm within *Urban FPS Generator* at some point in time.

The agent captures user input through the selection process as indicated above. A set of abstract features such as occupied/free space, density of street system and number of (obstacle) props used are used as values for the training set. The agent uses all 9 maps of each iteration, with the preferred selection classified as positive, and the remaining 7 maps as rejected training samples. Over the course of a defined number of iterations (we use either 10 or 20 for testing purposes), the agent receives a small training set of selected and rejected candidates. We are aware of the limitations due to the relatively small number of samples in the training set, given that J48 classifiers are often trained using hundreds or thousands of samples. But even under these circumstances the contribution of the agent has shown improvements of the overall system performance by reducing the required number of iterations by the designer [13].

Additional agents can be added into the multi-agents system to support the game design process even further. For example evaluation agents are commonly modelled on behaviours similar to non-player characters (NPCs), which make use of simple artificial intelligence (AI) techniques such as decision trees and pathfinding. Keeping the AI simple allows for real time performance, an important property of NPCs. The capability of such evaluation agents is in turn relatively simple as well. We are currently looking at using ray casting to create advanced evaluation agents. Ray casting essentially allows for basic 'vision' so that agents can detect potential areas that serve as cover for players by identifying obstacles (obstructions to the cast rays). This process also enables agents to find areas that are lacking cover and could potentially be exploited by players and lead to so called choke points during multi-player matches. Further, ray casting works in three dimensional space, whereas path finding is limited to simple two dimensional maps. Ray casting is conveniently part of the Unity3D physics engine. It is therefore efficient and simple to use. Overall ray casting is a mechanism that provides significant improvements over simple path finding and advances common AI techniques used for NPCs. Figure 6 demonstrates the use of these techniques.

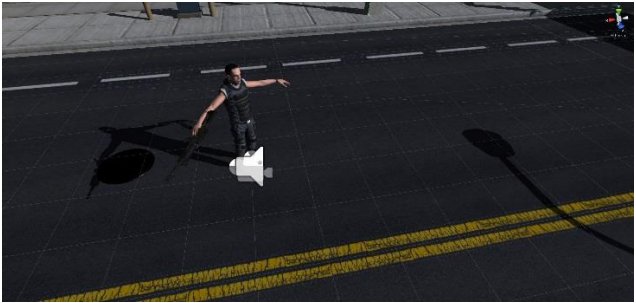

**Figure 6. NPC Using Advanced AI.**

## 4. FUTURE WORK

The work shown here serves as the foundation for future research into player behavior. We intent to use Cognitive Modelling [17] based on data gathered through in-game metric as well as player profile, demographic and other static factors. For the former we will utilize *Urban FPS Generator* as a testbed to extract data in a controlled way. Many commercial games, while providing the technical facilities to do so, allow no in-game measurements legally. *Urban FPS Generator* avoids these implications. The HBGA offers sufficient flexibility to customize the type and number of agents in a very economical way. It therefore provides a viable platform for future research into different generation and evaluation agents. That includes alternative machine learning algorithms to improve response to human fatigue during interactive iterations. But it also allows for more advanced evaluation agents testing game levels before they reach the breeding pool of the genetic algorithm. Finally, there is the possibility to model adaptive in-game AI agents that are used as NPCs to continuously adjust to human players, for instance as training bots or competitive partners. The main benefit of our work is the flexibility of the HBGA that allows fast development of different computational agents as part of the overall multi-agent system.

## 5. CONCLUSIONS

This paper introduces *Urban FPS Generator* a design tool for FPS map creation within the Unity3D game development environment using HBGA. We demonstrate its potential to create fully playable game levels with the option to run multi-player matches without any additional involvement of game designers or programmers. The system uses computational agents to compensate for limitations normally found in interactive evolutionary systems. Finally, *Urban FPS Generator* provides a platform for future research into player behaviour and team play including the option to derive in-game metrics during multi-player matches.